\documentclass[letterpaper]{article} 
\usepackage{aaai25}  
\usepackage{times}  
\usepackage{helvet}  
\usepackage{courier}  
\usepackage[hyphens]{url}  
\usepackage{graphicx} 
\usepackage{natbib}  
\usepackage{caption} 
\usepackage{algorithm}
\usepackage{algorithmic}
\usepackage{placeins}
\usepackage{url}
\usepackage{amsmath,amsfonts}
\usepackage{array}
\usepackage[caption=false,font=normalsize,labelfont=sf,textfont=sf]{subfig}
\usepackage{textcomp}
\usepackage{verbatim}
\usepackage{graphicx}
\usepackage{multirow}
\usepackage{multicol,lipsum}
\usepackage[table,xcdraw]{xcolor} 
\usepackage{newfloat}
\usepackage{listings}
\DeclareCaptionStyle{ruled}{labelfont=normalfont,labelsep=colon,strut=off} 
\floatstyle{ruled}
\newfloat{listing}{tb}{lst}{}
\floatname{listing}{Listing}
\title{SplitFedZip: Learned Compression for Data Transfer Reduction in Split-Federated Learning}
\author{Chamani Shiranthika, Hadi Hadizadeh, Parvaneh Saeedi, Ivan V. Baji\'{c}}
\affiliations{School of Engineering Science, Simon Fraser University, Burnaby, BC, Canada, V5A 1S6\\

    \{csj5, hadi\_hadizadeh, psaeedi, ibajic\}@sfu.ca
}

\usepackage{bibentry}

\begin{document}

\maketitle

\begin{abstract}
Federated Learning (FL) enables multiple clients to train a collaborative model without sharing their local data. Split Learning (SL) allows a model to be trained in a split manner across different locations. Split-Federated (SplitFed) learning is a more recent approach that combines the strengths of FL and SL. SplitFed minimizes the computational burden of FL by balancing computation across clients and servers, while still preserving data privacy. This makes it an ideal learning framework across various domains, especially in healthcare, where data privacy is of utmost importance. However, SplitFed networks encounter numerous communication challenges, such as latency, bandwidth constraints, synchronization overhead, and a large amount of data that needs to be transferred during the learning process. In this paper, we propose SplitFedZip -- a novel method that employs learned compression to reduce data transfer in SplitFed learning. Through experiments on medical image segmentation, we show that learned compression can provide a significant data communication reduction in SplitFed learning, while maintaining the accuracy of the final trained model. The implementation is available at: \url{https://github.com/ChamaniS/SplitFedZip}. 
\end{abstract}

%

\section{Introduction}
{F}{ederated} Learning (FL)~\cite{McMahan_2017} and Split Learning (SL)~\cite{Gupta_2018} are two powerful paradigms in distributed machine learning. FL enables collaborative training while maintaining client's data privacy, but places a significant computational burden on resource-constrained clients. In contrast, SL distributes model training across multiple devices, alleviating this burden. Split-Federated (SplitFed) learning~\cite{thapa2022splitfed} combines the strengths of both approaches, partitioning models and distributing tasks to reduce client's computational load while ensuring data privacy. This makes it an ideal learning framework for various scenarios, especially healthcare, where data privacy is of utmost performance~\cite{Shiranthika_review_2023}. However, SplitFed learning still encounters communication challenges, 
arising from the large amount of data that must be transferred during training. 

Although there is very limited research on SplitFed-specific challenges, there is considerable prior work on communication issues in FL and SL. Various approaches for model compression in FL were proposed, such as structured and sketched updates~\cite{konevcny_2016}, sparsification~\cite{isik2022sparse}, quantization~\cite{malekijoo2021fedzip}, autoencoders (AEs)~\cite{kong_2024, mudvari2024adaptive}, and knowledge distillation~\cite{wu2022communication}. 
The authors in~\cite{mitchell2022optimizing} first used rate-distortion (RD) optimization for model compression in FL. Our work also builds on RD optimization but applies it to the data being transferred in SplitFed: features and gradients. Meanwhile, compression in SL was achieved by exploiting circular convolution and correlation~\cite{hsieh_2022c3}, AEs~\cite{mudvariadaptive_2024}, top-k sparsification, and quantization~\cite{zheng2023reducing}. 
Comparatively, very few research studies examined data compression in SplitFed, 
including quantization~\cite{wang2022fedlite, oh2023communication} and feature compression using AEs~\cite{ayad_2021}. To our knowledge, this is the first work on end-to-end RD optimized data compression in SplitFed. 

In this paper, we propose SplitFedZip, a novel method that integrates learned compression into SplitFed to enhance communication efficiency by reducing the amount of data that needs to be transferred during training. For concreteness, we use two medical image segmentation datasets, and employ two codec architectures borrowed from image compression:  the custom AE~\cite{balle2018variational} and the Cheng2020 model with attention (Cheng\_AT)~\cite{cheng2020_learned}. Our approach optimizes these codecs by adjusting them to the statistics of the data involved in SplitFed learning -- features and gradients -- while SplitFed training is taking place. Our experiments show that SplitFedZIP significantly reduces the amount of data transferred during model training with minimal impact on the final model performance. 

The structure of the paper is as follows. We first describe our SplitFedZip methodology, followed by our experimental details. A comparison of SplitFedZip with the lone other SplitFed compression technique from~\cite{ayad_2021} comes next. Finally, the paper concludes by presenting the key findings and SplitFedZip's significance. 

\section{SplitFedZip}
\label{sec:methodology}
\subsection{Architectural design}
\label{sec:architecture}
\begin{figure}[!t]
\centering
\includegraphics[width=3.3in]{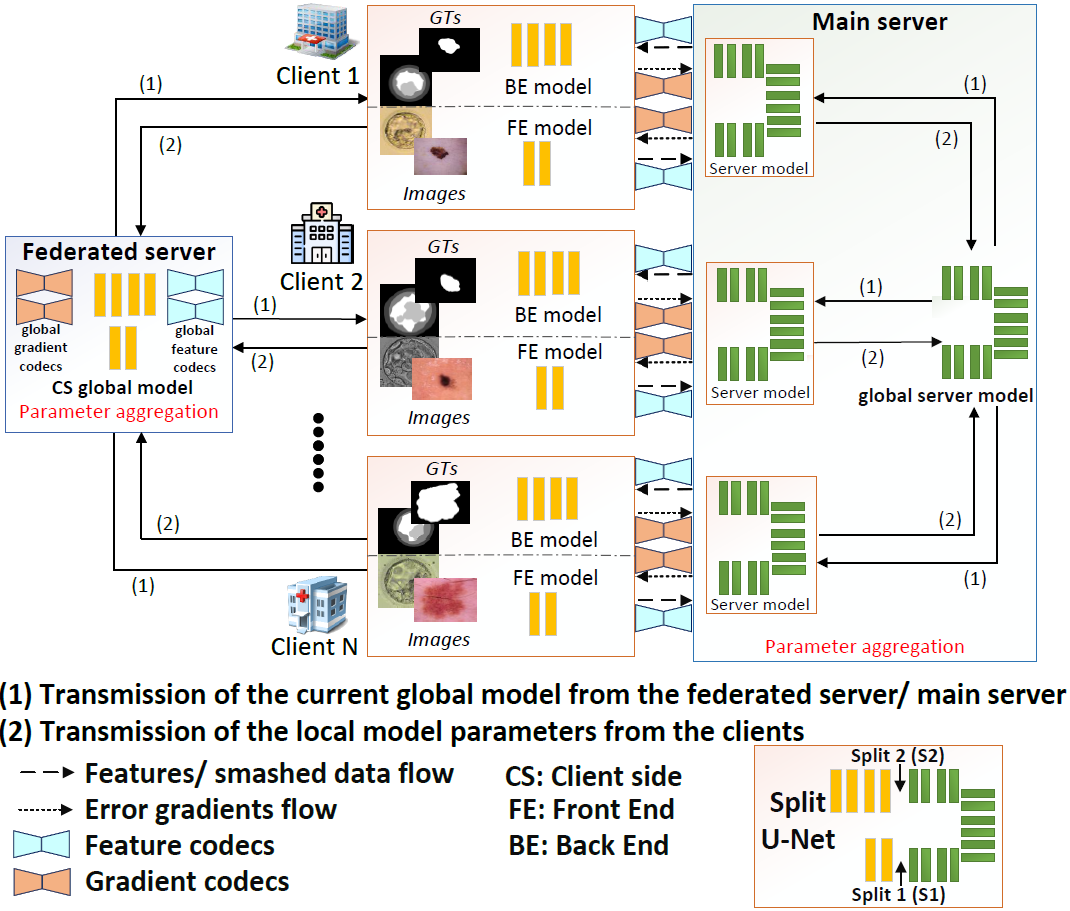}
\caption{Proposed SplitFedZip architecture.}
\label{fig:SplitFedzip}
\end{figure}

Our splitFedZip architecture shown in Fig. \ref{fig:SplitFedzip} consists of clients training a split U-Net~\cite{shiranthika_2023, Zahra_2023} for medical image segmentation. The split U-Net (bottom right corner of Fig.~\ref{fig:SplitFedzip}) has two split points: Split 1 (S1) and Split 2 (S2), splitting the entire U-Net into three parts: the front end (FE), server, and back end (BE) models. The FE model has access exclusively to the images, whereas the BE model has access solely to the ground truth (GT) labels. Both the FE and BE models comprise a small proportion of the entire U-Net and typically reside on the same client. In our case, the FE and BE models consist of 28.77 K and 73.16 K model parameters, and use 987.76 MMAC and 2.99 GMAC operations, respectively. By comparison, the server model consists of 7.66 M model parameters and uses 9.79 GMAC operations. The overall SplitFedZip system comprises a federated server, a main server, and multiple clients. The federated server holds the client-side global model (i.e., the global FE and BE models) along with two global feature and two global gradient codecs for compression at S1 and S2. The main server holds the global server model. Initially, global models and global codecs are randomly initialized and sent to all clients for local training.

During local training, each client trains its FE, server, and BE models with codecs to compress features and gradients. After several local epochs, models and codecs are aggregated via federated averaging \cite{McMahan_2017}, forming a new global model and new global codecs. This process continues for multiple global epochs until the global model converges. In the forward pass, features from the FE are compressed, transmitted, decompressed at the server, processed, and returned to the client for BE processing. Analogously to the features on the forward pass, in the backward pass during backpropagation, gradients are compressed prior to transmission and decompressed upon reception by the first and second gradient codecs, respectively.

The AE codec's encoder employs convolutional and generalized divisive normalization (GDN) layers~\cite{balle2016density} to transform features and gradients, while the entropy bottleneck quantizes and entropy-codes the transformed representations. The decoder uses transposed convolutional and inverse GDN layers to reconstruct the original data. The AE has 635.33 K model parameters and 2.76 GMAC computations. In contrast, the Cheng\_AT codec uses a deep convolutional neural network with attention mechanisms, extracting hierarchical features through convolutional layers and residual blocks. Its attention module prioritizes challenging segments of input data, optimizing bit allocation, and utilizes discretized Gaussian mixture likelihoods for sophisticated entropy modeling. The Cheng\_AT codec contains 3.49 M model parameters and 2.64 GMAC computations. For both codecs, the input and output layers were adjusted to align with the feature and gradient tensor dimensions.
\subsection{Training loss}
\label{sec:loss_fun}
Each client in the SplitFedzip network trains their split U-Net and the codecs with a loss function of the form:
\begin{equation}
\label{basic_eq2}
L = L_{r} + \lambda \cdot \{L_{Dice} + L_{mse} \},
\end{equation}

where $L_{r}$ is the rate term obtained from the codec's entropy estimator, $L_{Dice}$ is the Dice coefficient \cite{sudre_2017} used to quantify segmentation accuracy, and $L_{mse}$ is the Mean Squared Error (MSE) between codec's input and output data. $\lambda$ is the parameter that controls the relative weighting of the rate term $L_{r}$ and the distortion term $ \{L_{Dice} + L_{mse} \}$. Higher values of $\lambda$ place greater emphasis on distortion, leading to improved accuracy at the expense of increased bit-rate. Lower values of $\lambda$ prioritize reducing bit-rate, resulting in diminished task performance. 

We compared two compression schemes based on the type of data compressed, with loss components identified by the split point (S1 or S2) and data type (F for features in the forward pass and/or G for gradients in the backward pass). 
\begin{enumerate}
\item FG scheme: compressing both F and G, with the loss function:
\begin{flalign}
L &= \sum_{i=1,2} \left( L^{S{i},F}_{r} + L^{S{i},G}_{r} \right) +  \nonumber\\
&\lambda \cdot \left( \sum_{i=1,2} \left( L^{S{i},F}_{mse} + L^{S{i},G}_{mse} \right)  
+  L_{Dice}\right).
\label{FG_eq}
\end{flalign}
\item F scheme: compressing only F, with the loss function:
\begin{equation}
L = \sum_{i=1,2}  L^{S{i},F}_{r} + 
\lambda \cdot \left( \sum_{i=1,2} L^{S{i},F}_{mse} + L_{Dice} \right). \\
\label{F_eq}
\end{equation}
\end{enumerate}
The loss function when there's no compression (NC) is $L_{Dice}$.

\section{Experiments}
\label{sec:exp}
\subsection{Experimental setup}
\label{sec:exp_setup}
We investigated SplitFedZip on two datasets: the Blastocyst dataset~\cite{lockhart_2019} (781 samples) with five segmentation classes (trophoectoderm, zona pellucida, blastocoel, inner cell mass, and background), and the Human Against Machine with 10K training images (HAM10K) dataset~\cite{tschandl_2018} (10,015 samples) with two segmentation classes (skin lesion and background). Both datasets are split among five clients, with 85\% for training and 15\% for validation. Clients' data distribution is: 240, 120, 85, 179, 87 for the Blastocyst dataset, and is: 6325, 241, 71, 2359, 9, for HAM10K. The test data size is 70 for the Blastocyst dataset, and 1010 for the HAM10K. Data augmentation techniques include flipping, resizing, and normalizing. The models are trained with soft Dice loss~\cite{sudre_2017} using Adam optimizer, a learning rate of $1 \times 10^{-4}$, a batch size of 1, and 12 local epochs and 10 global epochs. Segmentation accuracy is measured by the mean Jaccard index (MJI)~\cite{Cox_2008}, excluding the background class. The same set of $\lambda$ values ($2 \times 10^{-9} \leq \lambda \leq 10^{10}$), are used for both datasets and codecs. All models were implemented using PyTorch with CompressAI~\cite{begaint2020compressai} on high-performance computing resources provided by the Digital Research Alliance of Canada.\footnote{\url{https://alliancecan.ca/en}}

\subsection{Results and analysis}
\label{sec:exp_res}
\begin{figure}[t]
\centering
\includegraphics[width=3.3in]{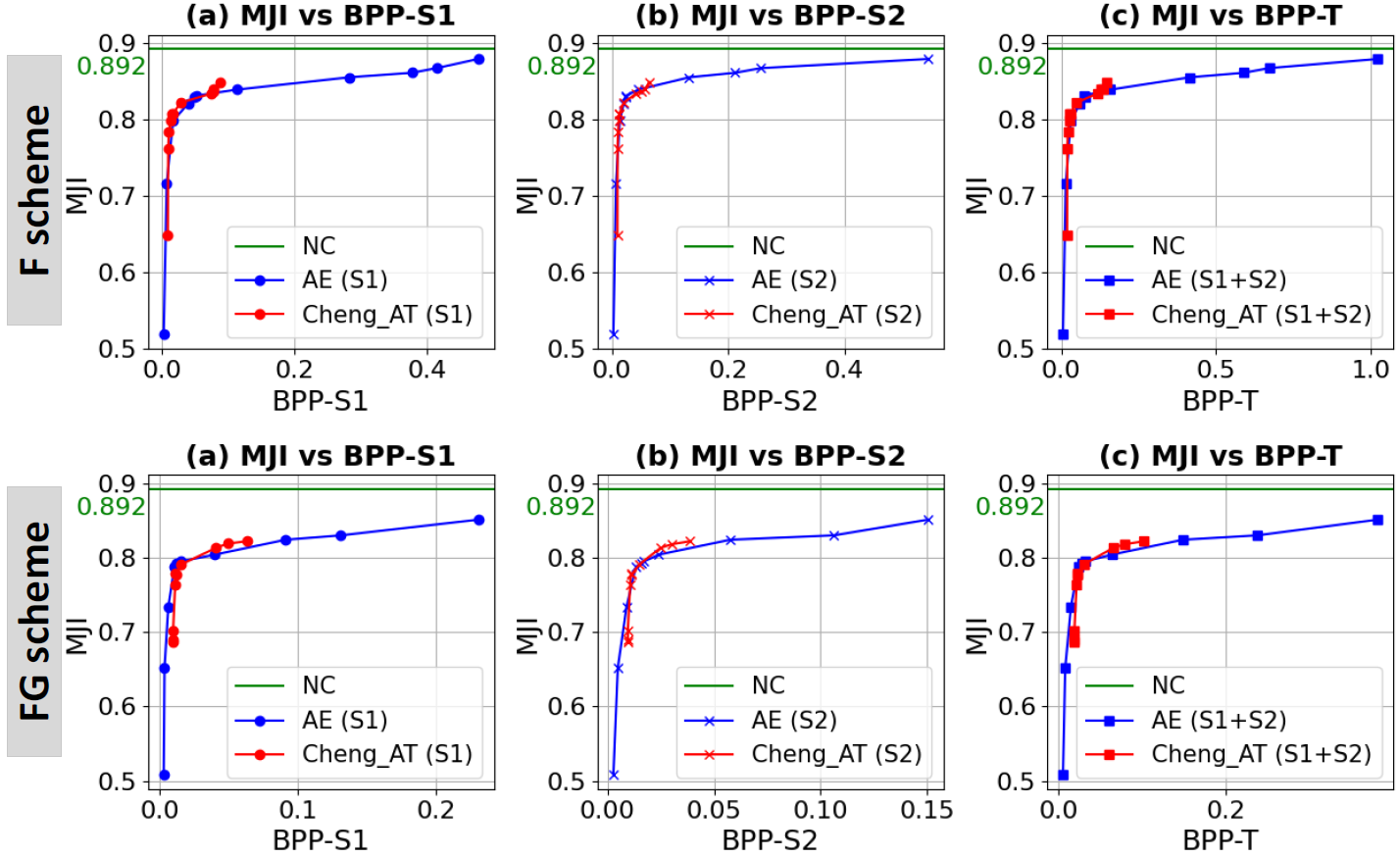}
\caption{R-A curves for the Blastocyst dataset.}
\label{fig:FandFG_Blasto}
\end{figure}

\begin{figure}[t]
\centering
\includegraphics[width=3.3in]{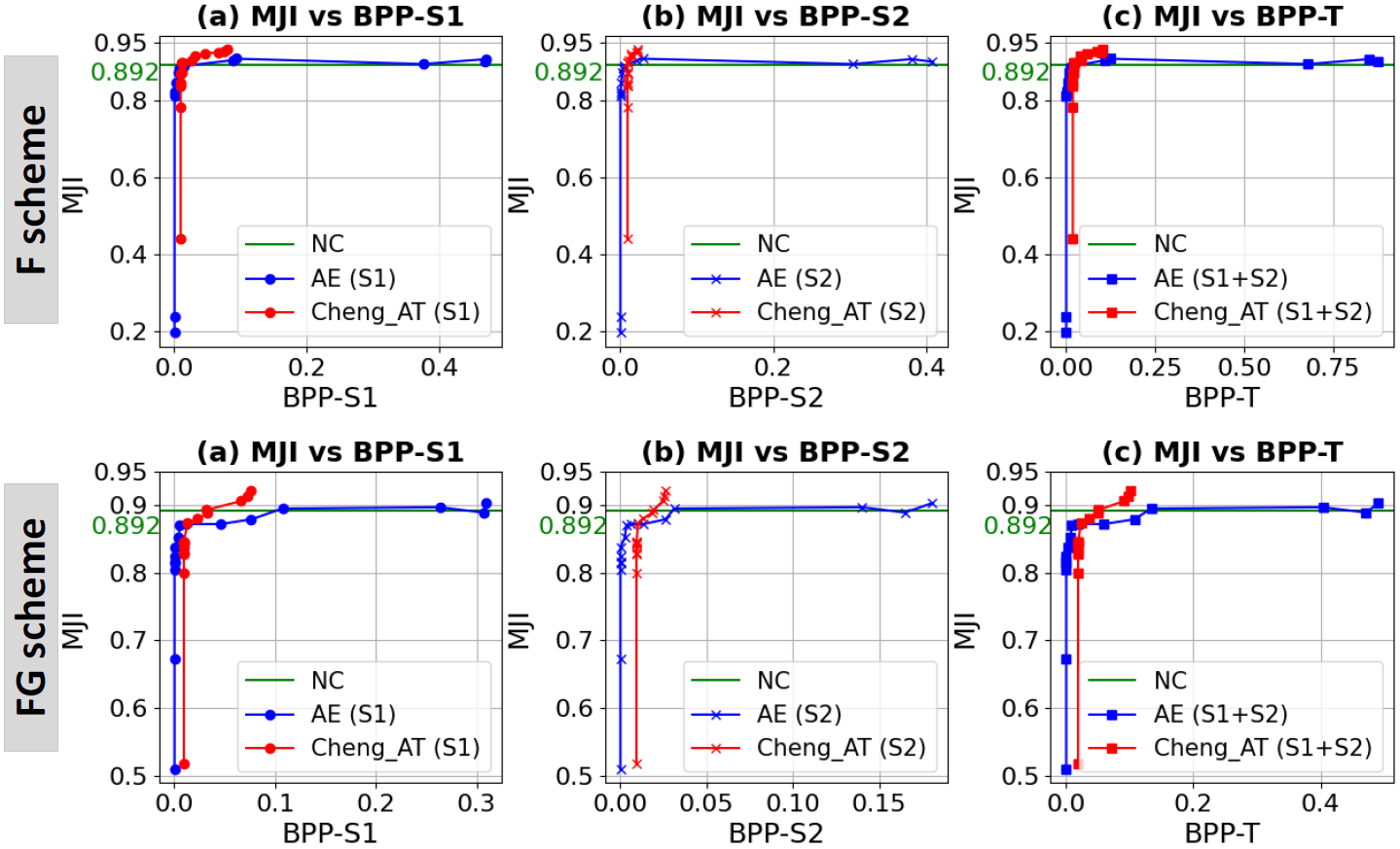}
\caption{R-A curves for the HAM10K dataset.}
\label{fig:FandFG_Ham}
\end{figure}

The bit-rates at S1 and S2 was measured in terms of Bits Per Pixel (BPP), defined as: the total bits in the compressed bitstream at S1 or S2 / input image resolution.
For each scheme (F and FG), we show three Rate-Accuracy (R-A) curves: MJI vs. BPP-S1, MJI vs. BPP-S2, and MJI vs. BPP-T (total bit-rate at S1 and S2) (Fig.~\ref{fig:FandFG_Blasto} and Fig.~\ref{fig:FandFG_Ham}). 
Different points on the R-A curves were obtained with the codecs trained using different $\lambda$ values. The horizontal green line represents the MJI during NC (0.892 for both datasets). On the Blastocyst dataset (Fig.~\ref{fig:FandFG_Blasto}), the accuracy of the final trained model gradually increases with bit-rate towards the NC performance. On the HAM10K dataset (Fig.~\ref{fig:FandFG_Ham}), the accuracy of the models trained with compression slightly surpasses that of the NC performance with bit-rates around 0.1 BPP. Apparently, features and gradients in the HAM10K dataset are much more compressible than those in the Blastocyst dataset, leading to a high rate-accuracy trade-off by both codecs. Moreover, the fact that HAM10K dataset is much larger (about 12x) compared to the Blastocyst dataset and that it corresponds to a somewhat simpler problem (binary segmentation vs. multi-class segmentation) means that the impact of compression is not as dramatic. In fact, compression on the HAM10K dataset seems to act as a regularizer and brings a slight improvement in the task performance. This phenomenon has also been observed in other works~\cite{Yang_etal_2022,Choi2022}.

To compare the AE and Cheng\_AT codecs, we utilized the Bjøntegaard Delta (BD) approach~\cite{bjontegaard2001calculation} calculating the average distance between the R-A curves. We computed BD-BPP (average BPP difference at the same MJI) and BD-MJI (average MJI difference at the same BPP), with AE as the anchor (Table~\ref{tab:bdcalculations}). A negative BD-BPP indicates reduced bit-rate compared to the anchor, which is the goal of compression. Weighted averages of BD-MJI and BD-BPP (WAvg) based on the test dataset sizes show that Cheng\_AT achieves 68.704\% bit-rate savings compared to the AE in the F scheme and 24.029\% in the FG scheme (bolded in Table \ref{tab:bdcalculations}). Clearly, the higher complexity and sophistication of the Cheng\_AT codec provides a significant advantage over the AE codec. 

\begin{table}[t]
\centering
\small 
\renewcommand*{\arraystretch}{1.1}
\begin{tabular}{>{\raggedright\arraybackslash}p{53pt}>
{\raggedright\arraybackslash}p{30pt}>
{\raggedright\arraybackslash}p{32pt}|>
{\raggedright\arraybackslash}p{30pt}> 
{\raggedright\arraybackslash}p{32pt}}
\hline 
Split point&\multicolumn{2}{c|} {F scheme} &\multicolumn{2}{c} {FG scheme} \\ \cline{2-5}
& BD-MJI & BD-BPP & BD-MJI  & BD-BPP  \\ \hline
\multicolumn{5}{c} {Blastocyst dataset (corresponds to Fig. \ref{fig:FandFG_Blasto})} \\ \hline
S1 & {0.004}  &  {--33.681}& {0.005}  &  {--99.195} \\ 
S2 & {0.007} & {--2.418 } & {0.007} & {--14.124 }\\ 
S1+S2 &{0.001}  & {--26.634}&{ 0.004}  & {--9.924}  \\ \hline
 \multicolumn{5}{c} {HAM10K dataset (corresponds to Fig. \ref{fig:FandFG_Ham})} \\ \hline 
S1 & {0.001}  &  {--62.019} & {0.010}  &  {--14.190}\\ 
S2 & {0.001} & {--25.089}& {0.007}  &  {--26.663}\\ 
S1+S2 &{0.002}  & {--71.620}& {0.009}  & {--25.007} \\ \hline
 \multicolumn{5}{c} {WAvg based on testing dataset sizes of the two datasets} \\ \hline
S1+S2: WAvg&{0.002}  & \textbf{{--68.704}}&{0.009}  &  \textbf{{--24.029}} \\ \hline
\end{tabular}
\caption{BD-MJI and BD-BPP (\%) values.}
\label{tab:bdcalculations}
\end{table}

\begin{figure}
\centering
\includegraphics[width=3.3in]{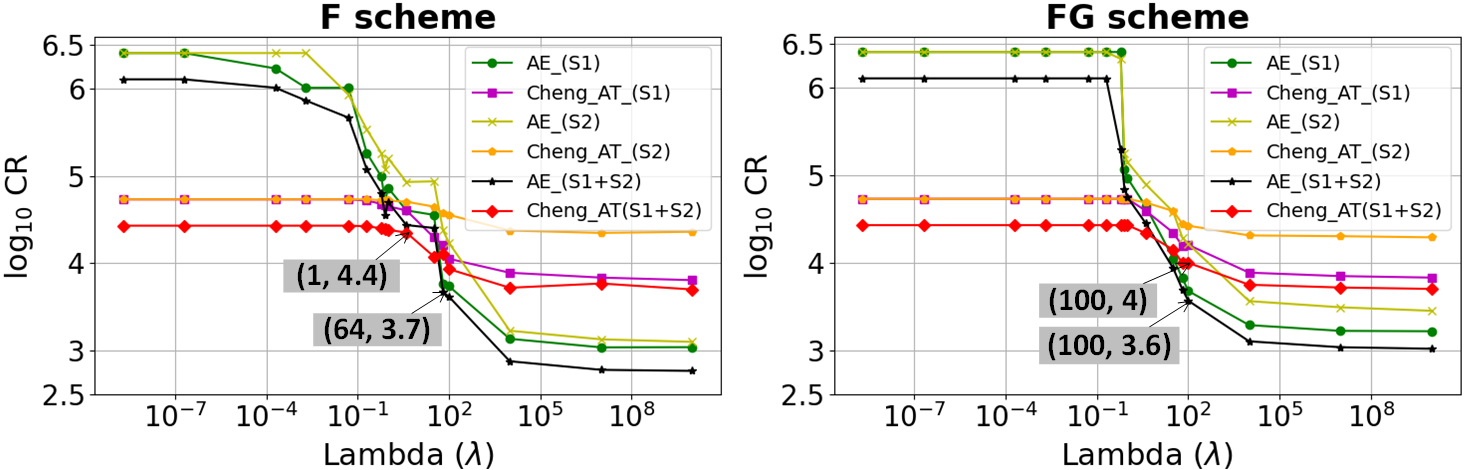}
\caption{CR vs. $\lambda$ for the HAM10K dataset.}
\label{fig:CR_HAM}
\end{figure}

Next, we illustrate the data transfer (DT) reduction advantage of SplitFedZip over NC. Fig.~\ref{fig:CR_HAM} shows the $\log_{10}$ of the compression ratio (CR), where CR is defined as: (original feature size at the split point S1 or S2) / (BPP-T $\times$ input image resolution), for different $\lambda$ values on the HAM10K dataset. As seen in the figures, across the range of $\lambda$, the CR ranges from $10^3$ to $10^{6.5}$. Comparing with Fig.~\ref{fig:FandFG_Ham}, we see that very high compression ratios (i.e., very low bit-rates) do not allow the model trained with compression to reach the NC performance. However, with a suitably chosen $\lambda$, the model trained with compression reaches the NC performance. For example, in the F scheme (Fig.~\ref{fig:CR_HAM} left), the AE codec allows SplitFed training to reach NC performance with $\lambda=64$, providing a CR of $10^{3.7}$. Meanwhile, the Cheng\_AT codec allows the same performance with $\lambda=1$, providing a CR of $10^{4.4}$. In the FG scheme (Fig.~\ref{fig:CR_HAM} right), the AE and Cheng\_AT codecs allow reaching NC performance with $\lambda=100$ and provide the CRs of $10^{3.6}$ and $10^4$, respectively. In these experiments, SplitFedZip reduces the transferred data by at least three orders of magnitude without compromising model performance.

Next, we examine the codec behavior as it adjusts to the statistics of the features and gradients while the main U-Net model is being trained. Figures~\ref{fig:BPP_analysis_Blasto} and~\ref{fig:BPP_analysis_HAM} in the Appendix present the training bit-rate of the Cheng\_AT codec with $\lambda = 1$ for the blastocyst and HAM10K datasets, respectively.
The bit-rates decrease during both local (left) and global (right) trainings as the codec learns to compress features and gradients. This bit-rate reduction enhances communication efficiency while preserving the necessary information for accurate updates. Stable convergence was observed, with any early-stage compression instabilities diminishing, leading to smooth, efficient training. In addition, gradient bit-rate is noticeably lower than the feature bit-rate: around 20\% lower in the Blastocyst dataset (Fig.~\ref{fig:BPP_analysis_Blasto} left) and around 6\% lower on the HAM10K dataset (Fig.~\ref{fig:BPP_analysis_HAM} left). When traded off against accuracy, as in eq.~(\ref{basic_eq2}), gradients can be compressed at a lower bit-rate than features.

Finally, Table \ref{tab:qualitative_comparison} in the Appendix visually compares two test samples, one from each dataset, using different $\lambda$ values. SplitFedZip predictions are from AE codecs under the FG scheme. There is a noticeable improvement in segmentation performance as $\lambda$ increases, with predictions increasingly aligning with GTs, as also seen in the rising MJI.

\subsection{Comparison with existing SplitFed compression methods}
\label{comparison}
SplitFedZip was compared to the only other compression technique in SplitFed~\cite{ayad_2021}, that used AEs solely for feature compression. It has two training methods: M1 (pretrained AE used as non-trainable layers in the split network), and M2 (AE trained for a few global epochs, then frozen). In both, gradients are back-propagated without further compression when the loss exceeds an adaptive threshold (\textit{gthres}; $0 \leq gthres \leq \infty$). For M1, we re-implemented their AE by pretraining on HAM10K for the Blastocyst dataset, and vice versa. DT is calculated as: (compressed feature size $\times$ no. of training samples) + (uncompressed gradient size $\times$ no. of uncompressed gradients). For a meaningful comparison, we adapted our FG scheme into \textit{two-stage training} (comparable to M1) and \textit{two-phase training} (comparable to M2), using our AE. In two-stage training, the split U-Net and AE are trained first on one dataset, and then the split U-Net is further trained with the frozen AE on the other dataset. In two-phase training, AE is trained with the split U-Net until the rate losses at S1 and S2 stabilize, then frozen for the rest. DT is calculated as: BPP-T $\times$ input image resolution $\times$ no. of training samples $\times$ 2. 

We show the MJIs and DTs for comparison in Table \ref{tab:merged_table} in the Appendix (top section-\cite{ayad_2021}'s AE, bottom section-SplitFedZip's AE). DT values are shown in GB. In the Blastocyst dataset, the highest MJI with M2 is 0.817 (at $gthres=0.25$, with 0.63 GB). SplitFedZip reached a similar MJI of 0.822 with just 0.02 GB of DT, offering a \textbf{32x reduction} (0.63/0.02). The corresponding cells are highlighted in dark green. In the HAM10K dataset, the highest MJI with M2 is 0.886 (at $gthres=0$, with 8 GB). SplitFedZip achieved a similar MJI of 0.892 using only 0.09 GB, a \textbf{89x reduction} (8/0.09). The corresponding cells are highlighted in dark purple. Significant savings were achieved with DT reductions of 3.2x (0.32/0.1) and 5000x (4/0.0008), relative to the M1 method in the Blastocyst (light green cells) and HAM10K (light purple cells) datasets, respectively. Therefore, SplitFedZip offers much better compression accuracy trade-off compared to~\cite{ayad_2021}. The reasons are: (1) SplitFedZip's end-to-end RD optimization, (2) more advanced entropy modeling in the codecs, allowing better adaptation to the statistics of the data being transferred, and (3) more sophisticated gradients compression, compared to simple thresholding in~\cite{ayad_2021}.

\section{Conclusion}
\label{sec:conc}
In this paper, we introduced SplitFedZip, the first end-to-end rate-distortion inspired compression approach for SplitFed learning. SplitFedZip optimizes compression of features and gradients passing through the split points in a SplitFed network simultaneously while the split network is being trained. We experimentally demonstrated SplitFedZip's effectiveness in transferred data reduction -- at least three orders of magnitude compared to no compression -- on two medical image segmentation datasets, without hindering the global model performance. Our experiments with two learnable codec architectures showed that the more sophisticated Cheng\_AT model yields better results. Compared to the lone other SplitFed compression approach in~\cite{ayad_2021}, SplitFedZip achieves equivalent accuracy while reducing data transfer by 3.2 - 5000 times. 
\newpage
\section{Acknowledgment}
\label{sec:ack}
The authors thank the Natural Sciences and Engineering Research Council (NSERC) of Canada for the financial support.

\bibliography{refs}

\begin{thebibliography}{30}
\providecommand{\natexlab}[1]{#1}

\bibitem[{Ayad, Renner, and Schmeink(2021)}]{ayad_2021}
Ayad, A.; Renner, M.; and Schmeink, A. 2021.
\newblock Improving the communication and computation efficiency of split learning for iot applications.
\newblock In \emph{Proc. IEEE GLOBECOM}, 01--06.

\bibitem[{Ball{\'e}, Laparra, and Simoncelli(2016)}]{balle2016density}
Ball{\'e}, J.; Laparra, V.; and Simoncelli, E.~P. 2016.
\newblock Density modeling of images using a generalized normalization transformation.
\newblock In \emph{Proc. ICLR}.

\bibitem[{Ball{\'e} et~al.(2018)Ball{\'e}, Minnen, Singh, Hwang, and Johnston}]{balle2018variational}
Ball{\'e}, J.; Minnen, D.; Singh, S.; Hwang, S.~J.; and Johnston, N. 2018.
\newblock Variational image compression with a scale hyperprior.
\newblock In \emph{Proc. ICLR}.

\bibitem[{B{\'e}gaint et~al.(2020)B{\'e}gaint, Racap{\'e}, Feltman, and Pushparaja}]{begaint2020compressai}
B{\'e}gaint, J.; Racap{\'e}, F.; Feltman, S.; and Pushparaja, A. 2020.
\newblock {CompressAI: A PyTorch} library and evaluation platform for end-to-end compression research.
\newblock \emph{arXiv preprint arXiv:2011.03029}.

\bibitem[{Bjontegaard(2001)}]{bjontegaard2001calculation}
Bjontegaard, G. 2001.
\newblock Calculation of average PSNR differences between RD-curves.
\newblock \emph{ITU-T SG16 Q}, 6.

\bibitem[{Cheng et~al.(2020)Cheng, Sun, Takeuchi, and Katto}]{cheng2020_learned}
Cheng, Z.; Sun, H.; Takeuchi, M.; and Katto, J. 2020.
\newblock Learned image compression with discretized gaussian mixture likelihoods and attention modules.
\newblock In \emph{Proc. CVPR}, 7939--7948.

\bibitem[{Choi and Bajić(2022)}]{Choi2022}
Choi, H.; and Bajić, I.~V. 2022.
\newblock Scalable Video Coding for Humans and Machines.
\newblock In \emph{Proc. IEEE MMSP}, 1--6.

\bibitem[{Cox and Cox(2008)}]{Cox_2008}
Cox, M. A.~A.; and Cox, T.~F. 2008.
\newblock \emph{Multidimensional Scaling}, 315--347.
\newblock Berlin, Heidelberg: Springer Berlin Heidelberg.
\newblock ISBN 978-3-540-33037-0.

\bibitem[{Gupta and Raskar(2018)}]{Gupta_2018}
Gupta, O.; and Raskar, R. 2018.
\newblock Distributed learning of deep neural network over multiple agents.
\newblock \emph{J. Netw. Comput.}, 116: 1--8.

\bibitem[{Hsieh, Chuang, and Wu(2022)}]{hsieh_2022c3}
Hsieh, C.~Y.; Chuang, Y.-C.; and Wu, A.-Y. 2022.
\newblock {C3}-{SL}: Circular convolution-based batch-wise compression for communication-efficient split learning.
\newblock In \emph{Proc. IEEE MLSP}, 1--6.

\bibitem[{Isik et~al.(2023)Isik, Pase, Gunduz, Weissman, and Zorzi}]{isik2022sparse}
Isik, B.; Pase, F.; Gunduz, D.; Weissman, T.; and Zorzi, M. 2023.
\newblock Sparse random networks for communication-efficient federated learning.
\newblock In \emph{Proc. ICLR}.

\bibitem[{Kafshgari et~al.(2023)Kafshgari, Shiranthika, Saeedi, and Bajić}]{Zahra_2023}
Kafshgari, Z.~H.; Shiranthika, C.; Saeedi, P.; and Bajić, I.~V. 2023.
\newblock Quality-Adaptive Split-Federated Learning for Segmenting Medical Images with Inaccurate Annotations.
\newblock In \emph{Proc. IEEE ISBI}, 1--5.

\bibitem[{Kone{\v{c}}n{\`y} et~al.(2016)Kone{\v{c}}n{\`y}, McMahan, Yu, Richt{\'a}rik, Suresh, and Bacon}]{konevcny_2016}
Kone{\v{c}}n{\`y}, J.; McMahan, H.~B.; Yu, F.~X.; Richt{\'a}rik, P.; Suresh, A.~T.; and Bacon, D. 2016.
\newblock Federated learning: Strategies for improving communication efficiency.
\newblock In \emph{Proc. NeurIPS}, 5--10.

\bibitem[{Kong et~al.(2024)Kong, Yu, Xu, Yu, Xu, and Huang}]{kong_2024}
Kong, Y.; Yu, W.; Xu, S.; Yu, F.; Xu, Y.; and Huang, Y. 2024.
\newblock Two Birds With One Stone: Towards Communication and Computation Efficient Federated Learning.
\newblock \emph{IEEE Commun. Lett.}

\bibitem[{Lockhart et~al.(2019)Lockhart, Saeedi, Au, and Havelock}]{lockhart_2019}
Lockhart, L.; Saeedi, P.; Au, J.; and Havelock, J. 2019.
\newblock Multi-Label Classification for Automatic Human Blastocyst Grading with Severely Imbalanced Data.
\newblock In \emph{Proc. IEEE MMSP}, 1--6.

\bibitem[{Malekijoo et~al.(2021)Malekijoo, Fadaeieslam, Malekijou, Homayounfar, Alizadeh-Shabdiz, and Rawassizadeh}]{malekijoo2021fedzip}
Malekijoo, A.; Fadaeieslam, M.~J.; Malekijou, H.; Homayounfar, M.; Alizadeh-Shabdiz, F.; and Rawassizadeh, R. 2021.
\newblock Fedzip: A compression framework for communication-efficient federated learning.
\newblock \emph{arXiv preprint arXiv:2102.01593}.

\bibitem[{McMahan et~al.(2017)McMahan, Moore, Ramage, Hampson, and y~Arcas}]{McMahan_2017}
McMahan, B.; Moore, E.; Ramage, D.; Hampson, S.; and y~Arcas, B.~A. 2017.
\newblock {Communication}-{Efficient} {Learning} of {Deep} {Networks} from {Decentralized} {Data}.
\newblock In \emph{Proc. AISTATS}, 1273--1282. PMLR.

\bibitem[{Mitchell et~al.(2022)Mitchell, Ball{\'e}, Charles, and Kone{\v{c}}n{\`y}}]{mitchell2022optimizing}
Mitchell, N.; Ball{\'e}, J.; Charles, Z.; and Kone{\v{c}}n{\`y}, J. 2022.
\newblock Optimizing the communication-accuracy trade-off in federated learning with rate-distortion theory.
\newblock \emph{arXiv preprint arXiv:2201.02664}.

\bibitem[{Mudvari et~al.(2024{\natexlab{a}})Mudvari, Vainio, Ofeidis, Tarkoma, and Tassiulas}]{mudvari2024adaptive}
Mudvari, A.; Vainio, A.; Ofeidis, I.; Tarkoma, S.; and Tassiulas, L. 2024{\natexlab{a}}.
\newblock Adaptive compression-aware split learning and inference for enhanced network efficiency.
\newblock \emph{ACM TOIT}, 24(4): 1--26.

\bibitem[{Mudvari et~al.(2024{\natexlab{b}})Mudvari, Vainio, Ofeidis, Tarkoma, and Tassiulas}]{mudvariadaptive_2024}
Mudvari, A.; Vainio, A.; Ofeidis, I.; Tarkoma, S.; and Tassiulas, L. 2024{\natexlab{b}}.
\newblock Adaptive Compression-Aware Split Learning and Inference for Enhanced Network Efficiency.
\newblock \emph{arXiv preprint arXiv:2311.05739}.

\bibitem[{Oh et~al.(2023)Oh, Lee, Brinton, and Jeon}]{oh2023communication}
Oh, Y.; Lee, J.; Brinton, C.~G.; and Jeon, Y.-S. 2023.
\newblock Communication-Efficient Split Learning via Adaptive Feature-Wise Compression.
\newblock \emph{arXiv preprint arXiv:2307.10805}.

\bibitem[{Shiranthika et~al.(2023)Shiranthika, Kafshgari, Saeedi, and Baji{\'c}}]{shiranthika_2023}
Shiranthika, C.; Kafshgari, Z.~H.; Saeedi, P.; and Baji{\'c}, I.~V. 2023.
\newblock SplitFed resilience to packet loss: Where to split, that is the question.
\newblock In \emph{Proc. MICAAI}, 367--377. Springer.

\bibitem[{Shiranthika, Saeedi, and Bajić(2023)}]{Shiranthika_review_2023}
Shiranthika, C.; Saeedi, P.; and Bajić, I.~V. 2023.
\newblock Decentralized Learning in Healthcare: A Review of Emerging Techniques.
\newblock \emph{IEEE Access}, 11: 54188--54209.

\bibitem[{Sudre et~al.(2017)Sudre, Li, Vercauteren, Ourselin, and Jorge~Cardoso}]{sudre_2017}
Sudre, C.~H.; Li, W.; Vercauteren, T.; Ourselin, S.; and Jorge~Cardoso, M. 2017.
\newblock Generalised dice overlap as a deep learning loss function for highly unbalanced segmentations.
\newblock In \emph{DLMIA ML-CDS}, 240--248. Springer.

\bibitem[{Thapa et~al.(2022)Thapa, Arachchige, Camtepe, and Sun}]{thapa2022splitfed}
Thapa, C.; Arachchige, P. C.~M.; Camtepe, S.; and Sun, L. 2022.
\newblock {SplitFed}: {When} {Federated} {Learning} {Meets} {Split} {Learning}.
\newblock In \emph{Proc. AAAI}, volume~36, 8485--8493.

\bibitem[{Tschandl, Rosendahl, and Kittler(2018)}]{tschandl_2018}
Tschandl, P.; Rosendahl, C.; and Kittler, H. 2018.
\newblock The {HAM10000} dataset, a large collection of multi-source dermatoscopic images of common pigmented skin lesions.
\newblock \emph{Sci. Data}, 5(1): 1--9.

\bibitem[{Wang et~al.(2022)Wang, Qi, Rawat, Reddi, Waghmare, Yu, and Joshi}]{wang2022fedlite}
Wang, J.; Qi, H.; Rawat, A.~S.; Reddi, S.; Waghmare, S.; Yu, F.~X.; and Joshi, G. 2022.
\newblock Fedlite: A scalable approach for federated learning on resource-constrained clients.
\newblock \emph{arXiv preprint arXiv:2201.11865}.

\bibitem[{Wu et~al.(2022)Wu, Wu, Lyu, Huang, and Xie}]{wu2022communication}
Wu, C.; Wu, F.; Lyu, L.; Huang, Y.; and Xie, X. 2022.
\newblock Communication-efficient federated learning via knowledge distillation.
\newblock \emph{Nat. Commun.}, 13(1): 2032.

\bibitem[{Yang, Amer, and Jiang(2021)}]{Yang_etal_2022}
Yang, E.~H.; Amer, H.; and Jiang, Y. 2021.
\newblock Compression Helps Deep Learning in Image Classification.
\newblock \emph{Entropy}, 23(7).

\bibitem[{Zheng et~al.(2023)Zheng, Chen, Lyu, and Yao}]{zheng2023reducing}
Zheng, F.; Chen, C.; Lyu, L.; and Yao, B. 2023.
\newblock Reducing communication for split learning by randomized top-k sparsification.
\newblock In \emph{Proc. IJCAI}, 4665--4673.

\end{thebibliography}
\newpage
\section{Appendix}
\label{sec:appendix}
\begin{figure}[!ht]
\centering
\includegraphics[width=3.3in]{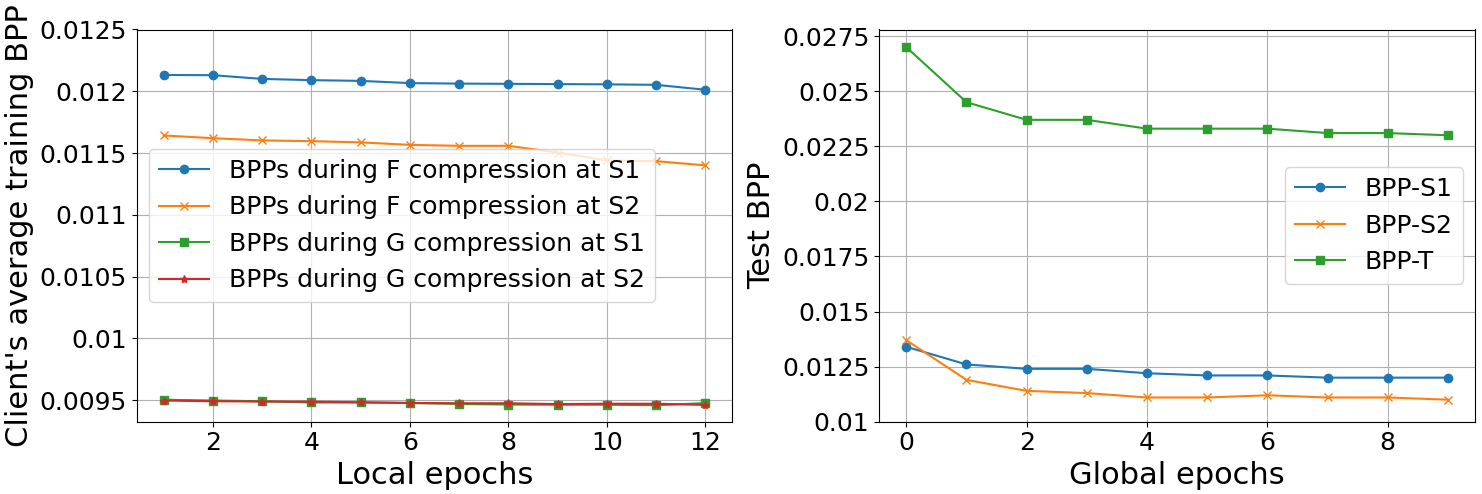}
\caption{Analysis of BPP during SplitFed training for the blastocyst dataset.}
\label{fig:BPP_analysis_Blasto}
\end{figure}
\begin{figure}[!ht]
\centering
\includegraphics[width=3.3in]{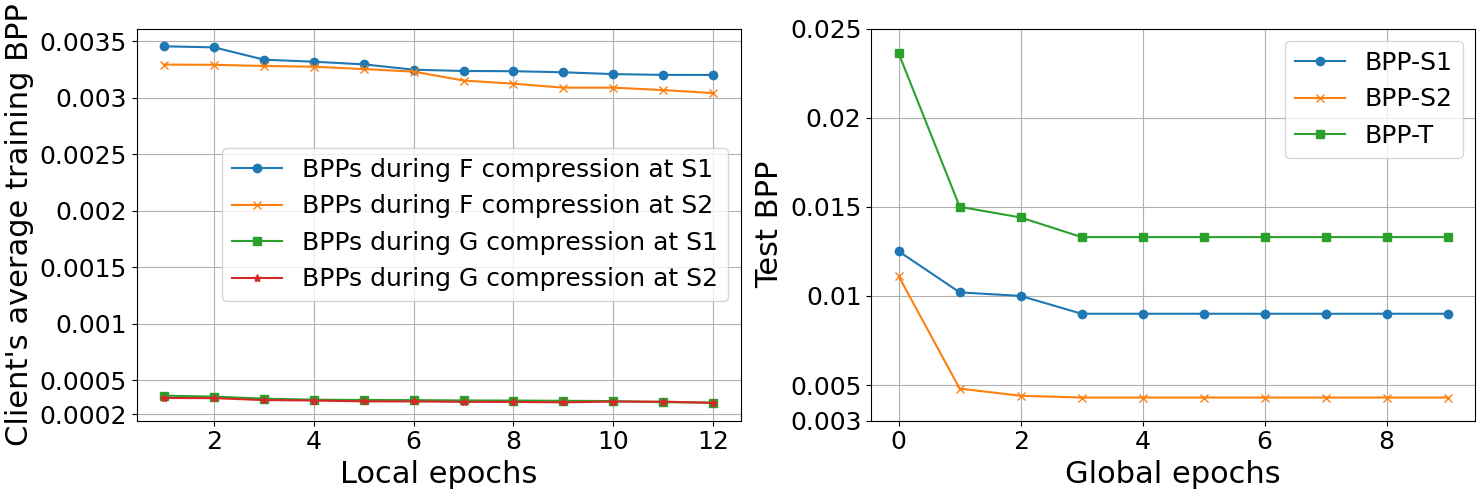}
\caption{Analysis of BPP during SplitFed training for the HAM10K dataset.}
\label{fig:BPP_analysis_HAM}
\end{figure}

\vspace{15mm}

\begin{table}[!ht]
\begin{center}
\small 
\renewcommand*{\arraystretch}{1.1}
\setlength{\tabcolsep}{1.5pt}
\begin{tabular}{>{\raggedright\arraybackslash}p{27pt}|>
{\raggedright\arraybackslash}p{27pt}|>
{\raggedright\arraybackslash}p{27pt}|>
{\raggedright\arraybackslash}p{27pt}|>
{\raggedright\arraybackslash}p{27pt}|>
{\raggedright\arraybackslash}p{27pt}|>
{\raggedright\arraybackslash}p{27pt}|>
{\raggedright\arraybackslash}p{27pt}}
\hline 
Image & Ground Truth (GT) &NC prediction &  
\multicolumn{5}{c} {SplitFedZip prediction} \\ 
\cline{4-8}
&  &  & \textbf{$\lambda = 0.0002$} & \textbf{$\lambda = 0.05$} & \textbf{$\lambda = 1$} & \textbf{$\lambda = 100$} & \textbf{$\lambda = 10^{10}$} \\  
\hline

\multicolumn{8}{c}{Blastocyst Dataset} \\ \hline
\parbox[c]{0.40in}{%
    \vspace{-0.12in}%
\begin{center}\includegraphics[width=0.40in,height=0.40in]{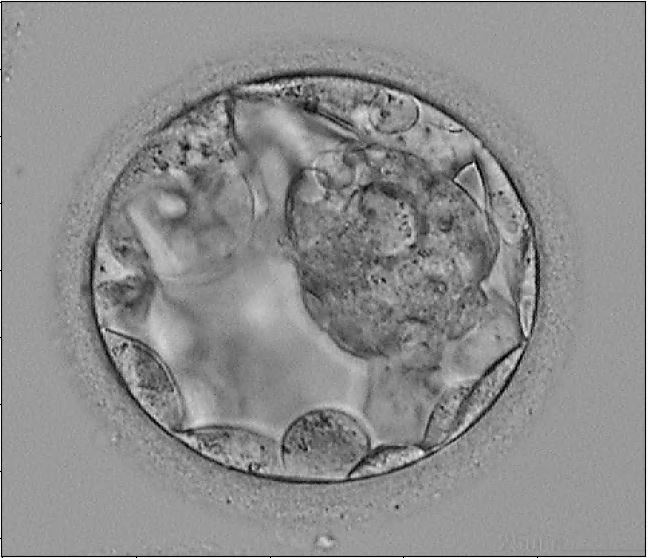}\\
        \tiny 182.BMP
    \end{center}
}
& 
\parbox[c]{0.40in}{%
    \vspace{-0.21in}%
\begin{center}\includegraphics[width=0.40in,height=0.40in]{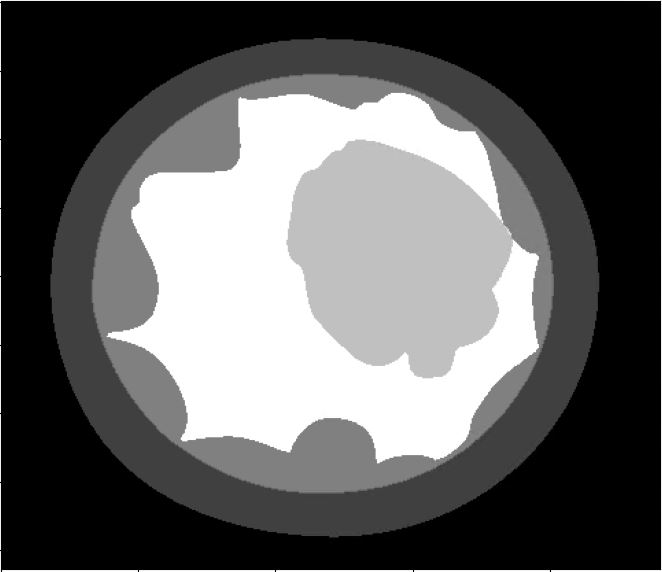}\\
    \end{center}
}
& 
\parbox[c]{0.40in}{%
    \vspace{-0.09in}%
\begin{center}\includegraphics[width=0.40in,height=0.40in]{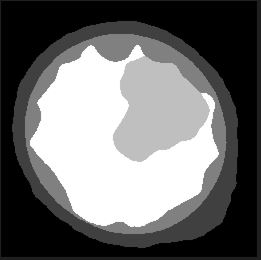}\\
       \small   $   0.892$
    \end{center}
}
&
\parbox[c]{0.40in}{%
    \vspace{-0.07in}%
\begin{center}\includegraphics[width=0.40in,height=0.40in]{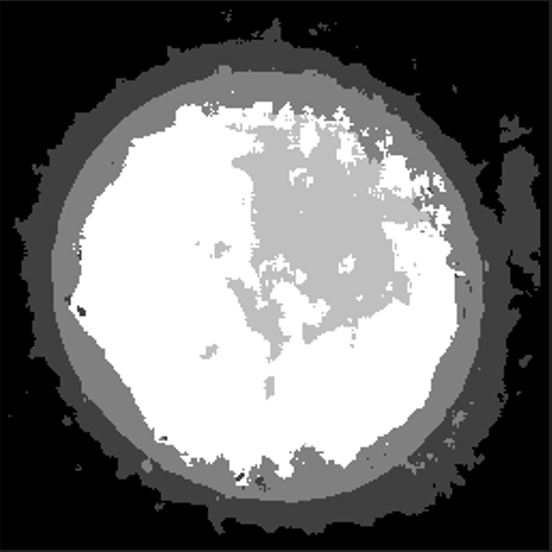}\\
      \small   $   0.474$
    \end{center}
}
& 
\parbox[c]{0.40in}{%
    \vspace{-0.07in}%
\begin{center}\includegraphics[width=0.40in,height=0.40in]{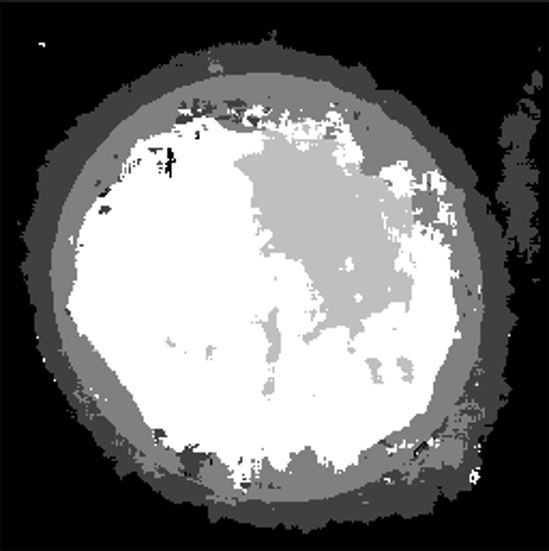}\\
     \small   $ 0.652$
    \end{center}
}
& 
\parbox[c]{0.40in}{%
    \vspace{-0.07in}%
\begin{center}\includegraphics[width=0.40in,height=0.40in]{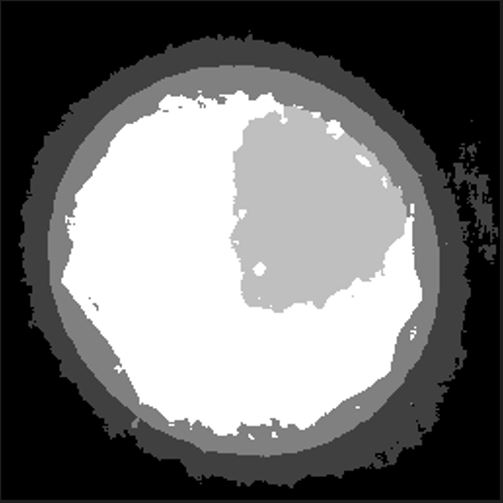}\\
  \small   $0.795$
    \end{center}
}
& 
\parbox[c]{0.40in}{%
    \vspace{-0.07in}%
\begin{center}\includegraphics[width=0.40in,height=0.40in]{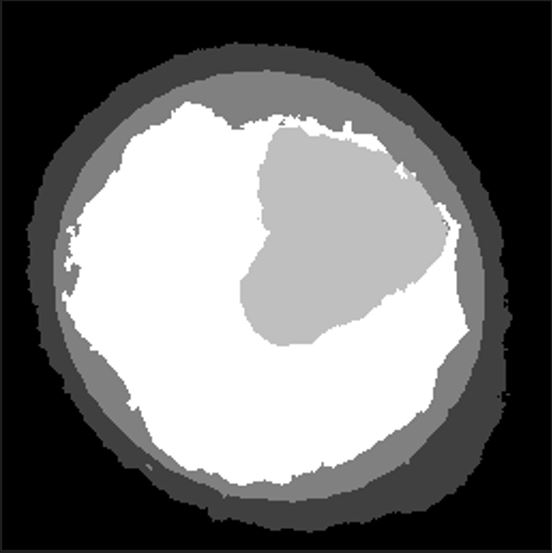}\\
        \small   $ 0.839$
    \end{center}
}
& 
\parbox[c]{0.40in}{%
    \vspace{-0.07in}%
\begin{center}\includegraphics[width=0.40in,height=0.40in]{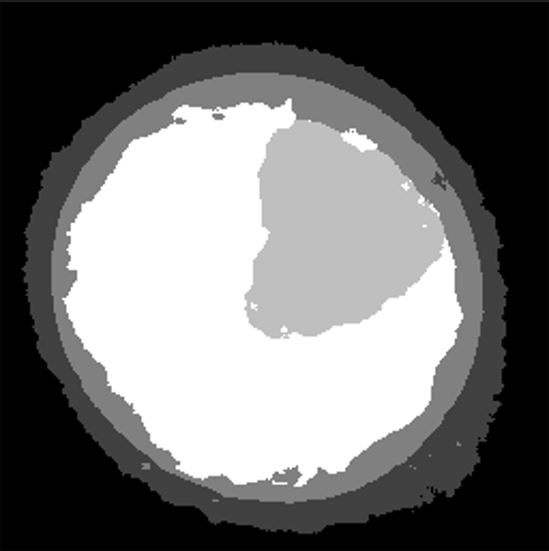}\\
    \small   $  0.843 $
    \end{center}
}
\\ \hline

\multicolumn{8}{c}{HAM10K Dataset} \\ \hline
\parbox[c]{0.40in}{%
   \vspace{-0.07in}%
\begin{center}\includegraphics[width=0.40in,height=0.40in]{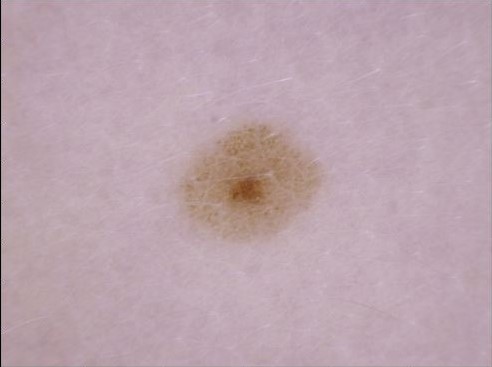}\\
      {\tiny \parbox{0.8\linewidth}{ISIC-00 \\ 25193.JPG}}
    \end{center}
}
& \parbox[c]{0.40in}{%
   \vspace{-0.27in}%
\begin{center}\includegraphics[width=0.40in,height=0.40in]{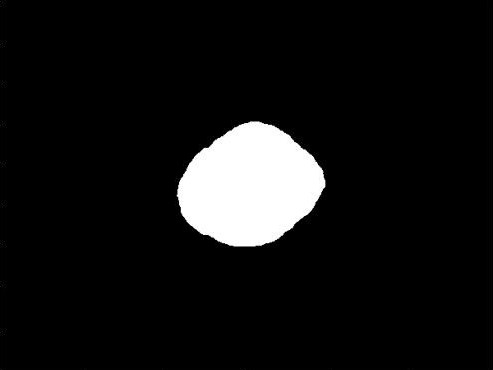}\\
    \end{center}
}
&\parbox[c]{0.40in}{%
    \vspace{-0.13in}%
\begin{center}\includegraphics[width=0.40in,height=0.40in]{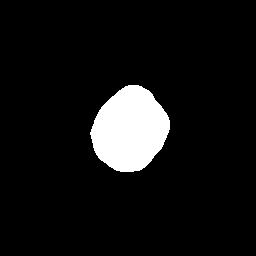}\\
       \small   $   0.892 $
    \end{center}
}
& \parbox[c]{0.40in}{%
    \vspace{-0.13in}%
\begin{center}\includegraphics[width=0.40in,height=0.40in]{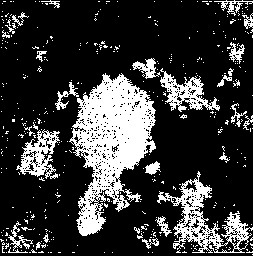}\\
     \small   $  0.663$
    \end{center}
}
& \parbox[c]{0.40in}{%
    \vspace{-0.13in}%
\begin{center}\includegraphics[width=0.40in,height=0.40in]{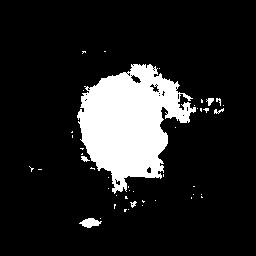}\\
  \small   $    0.804$
    \end{center}
}
& 
\parbox[c]{0.40in}{%
    \vspace{-0.13in}%
\begin{center}\includegraphics[width=0.40in,height=0.40in]{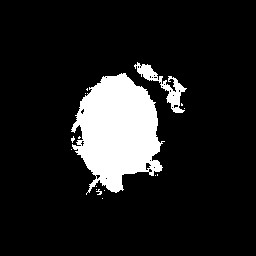}\\
    \small   $  0.871 $
    \end{center}
}
& \parbox[c]{0.40in}{%
    \vspace{-0.13in}%
\begin{center}\includegraphics[width=0.40in,height=0.40in]{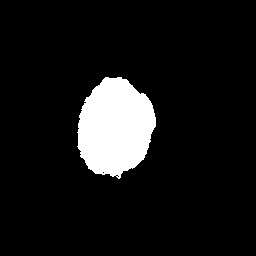}\\
     \small   $  0.895 $
    \end{center}
}
&  
\parbox[c]{0.40in}{%
    \vspace{-0.13in}%
\begin{center}\includegraphics[width=0.40in,height=0.40in]{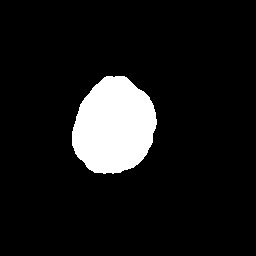}\\
  \small   $ 0.903 $
    \end{center}
}
\\ \hline
\end{tabular}
\end{center}
\caption{Qualitative comparison of two samples with different $\lambda$. The average MJI for the complete test sets are listed down.}
\label{tab:qualitative_comparison}
\end{table}

\vspace{-5mm}

\begin{table}[!ht]
\centering
\small
\renewcommand*{\arraystretch}{1.1}
\setlength{\tabcolsep}{3pt}
\begin{tabular}{>{\raggedright\arraybackslash}p{23pt}>
{\centering\arraybackslash}p{19pt}>
{\raggedleft\arraybackslash}p{19pt}>
{\centering\arraybackslash}p{19pt}>
{\raggedleft\arraybackslash}p{19pt}|>
{\centering\arraybackslash}p{19pt}>
{\raggedleft\arraybackslash}p{23pt}>
{\centering\arraybackslash}p{19pt}>
{\raggedleft\arraybackslash}p{23pt}}
    \hline 
    \multicolumn{9}{c}{\textbf{\cite{ayad_2021}'s AE}}\\ \hline
    \multirow{2}{*}{} & \multicolumn{4}{c|}{Blastocyst Dataset} & \multicolumn{4}{c}{HAM10K Dataset} \\ \cline{2-9} 
    \textit{gthres} & \multicolumn{2}{c}{M1} & \multicolumn{2}{c|}{M2} & \multicolumn{2}{c}{M1} & \multicolumn{2}{c}{M2} \\ \cline{2-9} 
    & MJI & DT & MJI & DT & MJI & DT & MJI & DT \\ \hline
    NC & 0.892 & 40.6 & 0.892 & 40.6 & 0.892 & 40.6 & 0.892 & 40.6 \\
    inf & 0.797  &0.32 & 0.475 &0.32 & 0.839 &4.0 & 0.802 &4.0\\ 
5   & 0.798  &0.32 & 0.548 &0.32 & 0.836 &4.0 & 0.808 &4.0\\ 
3   & 0.799  &0.32 & 0.717 &0.32 & 0.835 &4.0 & 0.865 &4.0\\ 
1.5 & 0.800  &0.32 & 0.789 &0.32 & 0.873 &4.0 & 0.860 &4.0\\ 
    1.0 & 0.799  & 0.32 & 0.787 & 0.32 & \cellcolor{blue!25} \textbf{0.876} & \cellcolor{blue!25}4.0 & 0.860 & 4.0 \\
    0.8 & 0.800  &0.32 & 0.796 &0.39 & 0.874 &4.0 & 0.860 &4.0\\
    0.5 & \cellcolor{green!25}\textbf{0.801} & \cellcolor{green!25}0.32 & 0.790 & 0.63 & 0.838 & 4.0 & 0.876 & 4.0 \\
    0.25 & 0.799 & 0.32 & \cellcolor{green!75}\textbf{0.817} & \cellcolor{green!75}0.63 & 0.841 & 4.0 & 0.873 & 4.0 \\
    0.2  & 0.799  &0.32 & 0.807 &0.63 & 0.868 &4.0 & 0.869 &4.1\\ 
0.15 & 0.798  &0.32 & 0.800 &0.63 & 0.869 &4.0 & 0.876 &4.1\\ 
0.10 & 0.798  &0.41 & 0.812 &0.63 & 0.844 &4.0 & 0.879 &4.3\\ 
0.08 & 0.800  &0.52 & 0.808 &0.63 & 0.823 &4.1 & 0.870 &4.4\\ 
0.05 & 0.795  &0.63 & 0.816 &0.63 & 0.828 &4.2 & 0.883 &5.0\\ 
    0.0 & 0.797 & 0.63 & 0.812 & 0.63 & 0.857 & 8.0 & \cellcolor{blue!45}\textbf{0.886} & \cellcolor{blue!45}8.0 \\ \hline
    \hline
    \multicolumn{9}{c}{\textbf{SplitFedZip's AE}} \\ \hline
    \multirow{2}{*}{} & \multicolumn{4}{c|}{Blastocyst Dataset} & \multicolumn{4}{c}{HAM10K Dataset} \\ \cline{2-9} 
    \textit{$\lambda$} & \multicolumn{2}{c}{Two-stage} & \multicolumn{2}{c|}{Two-phase} & \multicolumn{2}{c}{Two-stage} & \multicolumn{2}{c}{Two-phase} \\ \cline{2-9} 
    & MJI & DT & MJI & DT & MJI & DT & MJI & DT \\ \hline
    NC  & NA &NA &  NA& NA& NA & NA& NA &NA\\
    $10^{10}$ & 0.878& 1.04 &0.845 & 0.07 &0.894 &0.8  & \textbf{0.907} &0.6 \\ 
100  & 0.869 &0.5 & \textbf{0.847}  &0.04 & \textbf{0.898}  &0.3 &0.899  &0.12\\ 
    64 & \textbf{0.884} & 0.4 & 0.846 & 0.03 & 0.896 & 0.27 & \cellcolor{blue!50}0.892 & \cellcolor{blue!50}0.09 \\
    32 & 0.877 &0.4 &0.823  &0.03 &0.893  &0.06 &0.882  &0.04\\
    16 & 0.869 & 0.3 & \cellcolor{green!80}0.822 & \cellcolor{green!80}0.02 & 0.891 & 0.06 & 0.894 & 0.03 \\ 
    4 & \cellcolor{green!30}0.806 & \cellcolor{green!30}0.1 & 0.805 & 0.009 & 0.894 & 0.02 & 0.868 & 0.02 \\
    1 &  0.810&0.03 &0.778  &0.008 &0.889  &0.02 &0.869  &0.009\\ 
0.8 &0.804  &0.03 &0.769  &0.008 & 0.882 &0.02 & 0.864 &0.008\\ 
    0.6 & 0.799 & 0.02 & 0.775 & 0.005 & \cellcolor{blue!30}0.880 & \cellcolor{blue!30}0.0008 & 0.835 & 0.003 \\ 
    0.2 &  0.788& 0.02&  0.711& 0.003&0.839  &0.0004 &0.831  &0.002\\ 
0.05 &0.755  &0.018 &0.624  &0.003 & 0.834 &0.0004 &0.827  &0.0009\\ 
0.002 &0.607  &0.017 &0.512  &0.003 & 0.832 &0.0004&0.823 &0.0005  \\ 
0.0002 & 0.485 &0.005 &0.497  &0.003 & 0.808 &0.0004 &0.820  &0.0004\\ \hline
\end{tabular}
\caption{MJI and DT Comparison of \cite{ayad_2021}'s AE with SplitFedZip'AE. The highest value in each MJI column is bolded.}
\label{tab:merged_table}
\end{table}
\end{document}